\pdfoutput=1

\documentclass[11pt]{article}
\usepackage[]{acl}
\usepackage{times}
\usepackage{latexsym}
\usepackage{amssymb}
\usepackage{amsmath}
\usepackage[normalem]{ulem}
\usepackage[T1]{fontenc}
\usepackage[utf8]{inputenc}
\usepackage{microtype}
\usepackage{enumitem}
\usepackage{xcolor, graphicx}
\usepackage{multicol, multirow, array}
\usepackage{subcaption}

\title{From Partial to Strictly Incremental Constituent Parsing}

\author{Ana Ezquerro, Carlos Gómez-Rodríguez and David Vilares \\
  Universidade da Coruña, CITIC \\
  Departamento de Ciencias de la Computación y Tecnologías de la Información \\
  Campus de Elviña s/n, 15071 \\
  A Coruña, Spain\\
  \texttt{\{ana.ezquerro, carlos.gomez, david.vilares\}@udc.es}}

\usepackage{xcolor, fontawesome}
\usepackage{accents, pgffor}
\newcommand{\cvdots}[1]{
$
  \mathbin{\vcenter{\baselineskip.9ex
    {\foreach \X in {#1}{\hbox{\textcolor{\X}{\large\bf.}}}}
  }}%
$}

\usepackage{bibentry}
\usepackage{svg, svg-extract}

\def\SPSB#1#2{\rlap{\textsuperscript{#1}}\SB{#2}}
\def\SP#1{\textsuperscript{#1}}
\def\SB#1{\textsubscript{#1}}
\definecolor{gptcolor}{HTML}{22C99A}
\definecolor{xlmcolor}{HTML}{883BEB}
\definecolor{lstmcolor}{HTML}{F45D51}
\definecolor{mlpcolor}{HTML}{2796FE}
\definecolor{gcncolor}{HTML}{0B8F88}
\definecolor{pink}{HTML}{F477AD}
\definecolor{green}{HTML}{1F945D}
\definecolor{boostcolor}{HTML}{1F945D}

\newcommand{\mlp}{\textcolor{mlpcolor}{\faCaretUp}}
\newcommand{\bloom}{\textcolor{pink}{\faGe}}
\newcommand{\mgpt}{\textcolor{gptcolor}{\faGg}}
\newcommand{\lstm}{\textcolor{lstmcolor}{$\to$}}
\newcommand{\bilstm}{\textcolor{lstmcolor}{$\leftrightarrow$}}
\newcommand{\xlm}{\textcolor{xlmcolor}{\faSkyatlas}}
\newcommand{\gcn}{\textcolor{gcncolor}{\faEllipsisH}}
\newcommand{\abs}{a}
\newcommand{\rel}{r}

\begin{document}
\maketitle
\begin{abstract}
 We study incremental constituent parsers to assess their capacity to output trees based on prefix representations alone. Guided by strictly left-to-right generative language models and tree-decoding modules, we build parsers that adhere to a strong definition of incrementality across languages. This builds upon work that asserted incrementality, but that mostly only enforced it on either the encoder or the decoder. Finally, we conduct an analysis against non-incremental and partially incremental models.
\end{abstract}

\section{Introduction}

Incremental NLP aims to learn and adapt partial representations as information unfolds. 
However, with the rise of bidirectional LSTMs \cite{hochreiter1997long} and Transformers \cite{vaswani2017attention},  
recent research has focused on non-incremental solutions.
These models process the full input for contextualization before they start generating any output. Therefore, this approach does not capture the progressive unfolding of input over time, giving the sense that all of it is available all of a sudden \cite{madureira-schlangen-2020-incremental}. This is not an issue for most NLP tasks, but it is relevant for others, such as real-time NLP, e.g., instant machine translation or real-time speech. Furthermore, work on incremental processing holds relevance in interdisciplinary research, especially where computer science, linguistics, and cognitive studies intersect.

While some studies have addressed the challenge of outputting incremental structured representations - for various definitions of incrementality \cite{konstas-etal-2014-incremental,kohn-2018-incremental,shen-etal-2021-explicitly} -  analyses of trees remain limited, more notably since the popularization of deep learning, and are mostly partially incremental approaches.

In this context, \citet{titov-henderson-2007-constituent}, one of the first neural parsing models, was also an incremental network based on sigmoid belief networks. This generative model broke down the probability of a structure into probabilities for individual derivation decisions,
each influenced by previous decision history. However, the computation was expensive and its evaluation was restricted to sentences of up to 15 tokens in the English Penn Treebank \cite{marcus1993building}. For shift-reduce constituent parsing, \citet{cross-huang-2016-incremental} proposed an incremental model with minimal features, focusing on only three sentence positions to predict the next action. However, input sentences were contextualized using bidirectional LSTMs, thus relying on non-incremental encoders and effectively considering all upcoming words; a strategy that was later widely adopted by most neural syntactic parsing architectures, but that does not adhere to a definition of strong incrementality. More recently, \citet{kitaev-etal-2022-learned} introduced a span-based model that incrementally encodes input sentences into discrete elements using vectors from GPT-2 mapped into a codebook. Despite this, it relied on bidirectional Transformers and a CYK architecture \cite{kitaev2018constituency} for decoding these vectors into trees. Complementarily, \citet{yang2020strongly} proposed an incremental decoder based on graph neural networks. Although they referred to their parser as strongly incremental, sentences were encoded with bidirectional architectures like BERT or XLNET \cite{devlin-etal-2019-bert,yang2019xlnet}.

Incrementality has been also explored for other parsing formalisms. \citet{stanojevic-steedman-2019-ccg} developed an almost fully incremental parser for combinatory categorical grammars (CCG), relying on ELMo embeddings \cite{peters-etal-2018-deep} and a bidirectional LSTM for these predictions. Later, a genuinely fully incremental CCG parser was introduced \cite{stanojevic-steedman-2020-max}, using only ELMo's forward pass and a left-to-right LSTM, addressing biases in incremental CCG parsing. In the field of dependency parsing, incrementality has been a focus since the pre-neural era \cite{beuck2013structural,kohn-menzel-2014-incremental,kohn-baumann-2016-predictive}, with some models rivaling non-incremental ones. Recently, \citet{ezquerro2023challenges} found that with current neural architectures, incremental models for dependency parsing are less effective than bidirectional approaches. However, incorporating human-like reading strategies, such as brief delays, can significantly enhance performance, particularly in languages with leftward dependencies.

\paragraph{Contribution} 
We study the viability and challenges of \emph{fully incremental} constituent parsing with encoder-decoder architectures. All components strictly process the sentence from left to right, adding each read word to the partial tree based on the input prefix. For the encoder, we leverage generative LLMs. For the decoder, we reassess two options that generate partial trees based solely on current inputs: (i) an incremental parsing-as-tagging model \cite{gomez-rodriguez-vilares-2018-constituent}, and (ii) a transition-based decoder that uses graph-neural-network representations \cite{yang2020strongly}. The code is available at {\small\url{https://github.com/anaezquerro/incpar}}.

\section{Incremental Constituent Parsing}

Let $w = (w_1, ..., w_n)$ be a sequence of tokens such that $w_i \in \mathcal{V}$ for some vocabulary of tokens $\mathcal{V}$, we are interested in learning a  function that can map $w$ into a constituent tree $T$. Different from previous work, we are interested in modeling this  function as an strictly incremental model. Under this setup, the decision at time step $i$ is based only on the \emph{prefix} $w_1...w_{i+k}$. It creates a partial tree, $T_{i}$, where each word $w_i$ is added at its time step $i$, in a monotonic way. The delay parameter, $k$, mimics human reading processes, allowing for a slight look ahead to the upcoming words. Human parsing is believed to be very swift, with latencies as short as 250 milliseconds \cite{pulvermuller2009understanding, bemis2011simple}. In this work, we will study both zero and small positive delays, i.e., $k \in [0,2]$. Next, we review our encoders (\S \ref{section-incremental-encoders}) and decoders (\S \ref{section-incremental-decoders}).

\subsection{Incremental encoders}\label{section-incremental-encoders}

The incremental encoder is a parameterized function $\Psi_\theta$ that produces a hidden representation vector $\mathbf{h}_i\in\mathbb{R}^h$  for each input token $w_i$ based on its own prefix, thus $\mathbf{h}_i = \Psi_\theta(w_1...w_{i})$. As for specific architectures, will rely on encoders both without and with pre-training. The former is a lower-bound baseline made of 4 stacked left-to-right LSTMs \cite{hochreiter1997long}. For the latter, we use multilingual GPT \citep[mGPT;][]{mgpt} and BLOOM-560M \cite{scao2022bloom}. mGPT has pre-training data for all languages studied, while BLOOM does not. This lets us measure the impact of: (i) no pre-training data, (ii) pre-training data for all languages, and (iii) missing pre-training data for some languages (see also \S \ref{section-experiments}).

\subsection{Incremental decoders}\label{section-incremental-decoders}

We propose two different architectures to implement our incremental decoders. In both cases, an intermediate module was added between the encoder and decoder to add prefix information up to word $w_{i+k}$. At each timestep $i$, this module accepts the encoder representations $\mathbf{h}_i...\mathbf{h}_{i+k}$ and generates a new delayed contextualization $\overline{\mathbf{h}}_i$ using a feed-forward network ($\overline{\mathbf{h}}_i=\textbf{FFN}(\mathbf{h}_i...\mathbf{h}_{i+k})$). The delayed sequence $\mathbf{\overline{H}} = (\overline{\mathbf{h}}_1 ..., \overline{\mathbf{h}}_n)$ is directly passed as input to the decoder. Thus, these decoders produce an extra piece of the output tree based strictly on the prefix $w_1...w_{i+k}$.

On the one hand, we use decoders rooted in sequence labeling parsing~\cite{gomez-rodriguez-vilares-2018-constituent}. Here, at each time step, each representation is mapped to a partial label that encodes a segment of the constituent tree primarily based on the preceding prefix. On the other hand, we choose the incremental decoder by \citet{yang2020strongly}. They use a graph neural network to contextualize the partial tree and make a decision (transition) at each time step based on the read token.

\subsubsection{Incremental decoding as tagging}

\begin{figure}\centering 
    \includegraphics[width=0.45\textwidth]{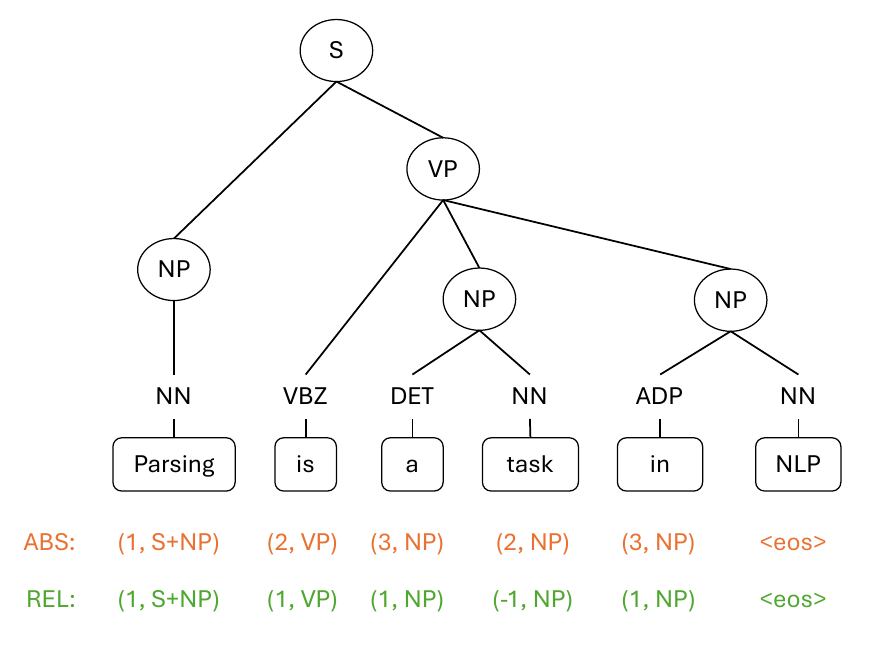}   
    \caption{\label{fig:sl-example} Absolute (orange) and relative (green) indexing from \citet{gomez-rodriguez-vilares-2018-constituent}. Note that unary chains are collapsed in an artificial constituent (first label). The final label indicates the end of sentence.}
\end{figure}

Given a sequence of delayed word contextualizations $\overline{\mathbf{H}}=(\overline{\mathbf{h}}_1...\overline{\mathbf{h}}_n)$, a tagging-based decoder maps each contextualization $\overline{\mathbf{h}}_i$ to a label $\ell_i\in\mathcal{L}$ and defines an injective and complete function to delinearize the sequence of labels into a valid constituent tree. Following \citet{gomez-rodriguez-vilares-2018-constituent}, each label is a tuple of the form $\ell_i$=$(d_i, c_i) \in \mathcal{L}$, where $d_i$ encodes a number $l_i$, the total number of levels in common between $w_i$ and $w_{i+1}$, and $c_i$ encodes the lowest non-terminal in common between those two words.\footnote{The encoding is injective and complete for constituent trees without unary chains. The specifics can be found in the reference paper. Here, unary chains were collapsed in a single artificial constituent and recovered in the decoding step.} $l_i$ can be encoded in $d_i$ either directly ($d_i = l_i$, absolute encoding) or as a difference from the previous value ($d_1 = l_1$ and $d_i = l_i - l_{i-1}$ for $i \ge 2$, relative encoding). See Figure~\ref{fig:sl-example} for an example.

We chose this encoder over newer sequence labeling linearizations that have been recently published, such as tetra-tagging \cite{kitaev-klein-2020-tetra} and  shift-reduce parsing through pre-order, post-order, and in-order linearizations \cite{amini-cotterell-2022-parsing}. We did so due to a few practical reasons: (i) it is more user-friendly with existing libraries for transforming constituent trees into label sequences; and (ii) it accommodates non-binary trees, like the juxtapose model (binarizing and unbinarizing is trivial, yet necessary for these mentioned alternatives).\footnote{Also, even if \citeauthor{kitaev-klein-2020-tetra} and \citeauthor{amini-cotterell-2022-parsing} report better results, it is worth noting that the original papers cannot be directly compared in terms of results due to different implementations. For instance, \citeauthor{gomez-rodriguez-vilares-2018-constituent} relied on LSTMs and a simple decoder based on feed-forward networks, while the tetra-tagging paper used BERT and did not employ a sequence labeling decoder, but rather an efficient and simple dynamic programming approach.}

That said, our decoder is straightforward. Given an incrementally delayed contextualized input $\overline{\mathbf{H}}= (\overline{\mathbf{h}}_1,...,\overline{\mathbf{h}}_n)$, each label $\ell_i$ is computed as   $\textbf{FFN}_\ell(\overline{\mathbf{h}}_i)$, where $\mathbf{FFN}_\ell$ is a 1-layered feed-forward network with a softmax activation.

\subsubsection{Incremental decoding as transition-based parsing}

\begin{figure}
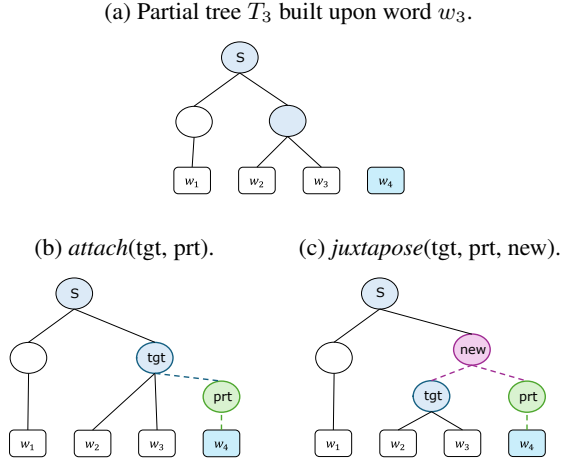

    \centering\small
    \begin{subfigure}[t]{0.5\textwidth}\centering 
        \caption{\label{fig:aj-state0}Partial tree $T_3$ built upon word $w_3$.} 
        \vspace{0.5em}
        \includesvg[scale=0.1,inkscapelatex=false]{figures/state0.svg}
    \end{subfigure}
    \\[1em] 
    \begin{subfigure}[t]{0.23\textwidth}\centering 
        \caption{\label{fig:aj-attach}\textit{attach}(tgt, prt).} 
        \vspace{0.5em}
        \includesvg[scale=0.1,inkscapelatex=false]{figures/attach.svg}
    \end{subfigure}
    \hfill
    \begin{subfigure}[t]{0.23\textwidth}\centering 
        \caption{\label{fig:aj-juxtapose}\textit{juxtapose}(tgt, prt, new).} 
        \vspace{0.5em}
        \includesvg[scale=0.1,inkscapelatex=false]{figures/juxtapose.svg}
    \end{subfigure}
    \caption{\label{fig:aj-example}Transitions defined by \citet{yang2020strongly} for a partial tree $T_3$ when a new word $w_4$ is added. Nodes in $\mathcal{R}(T_3)$ are marked in blue color.}
\end{figure}

Similar to the tag-based decoders, transition-based systems incrementally process each word contextualization to generate a sequence of actions of variable length $m$. Each action updates the system's inner representation of the partial tree until the sequence is fully processed and the final state retrieves the complete predicted tree. As a transition-based decoder, we use the strong incremental decoder by \citet{yang2020strongly}. It generates a sequence of $n$ transitions, adding exactly one token to the partial tree at each time step. Namely, each time step is represented by a partial tree $T_{i-1}$, which it is updated based on the subsequent $w_i$ and the rightmost chain of $T_{i-1}$ (denoted as $\mathcal{R}(T_{i-1})$)\footnote{Formally, the rightmost chain of a tree $T_{i-1}$ is defined by the set of non-terminal nodes whose rightmost fencepost coincides with the last word of the sentence (see Figure \ref{fig:aj-example}).} by performing one of these actions:

\begin{itemize}
    \item $\textit{attach}(\varphi^\mathrm{tgt}, \varphi^\mathrm{prt})$: Attaches a new subtree to $\mathcal{R}(T_{i-1})$. It creates a non-terminal parent node $\varphi^\mathrm{prt}$ and puts the $w_i$ as its terminal node. $\varphi^\mathrm{prt}$ also becomes the rightmost child of an existing non-terminal node $\varphi^\mathrm{tgt} \in \mathcal{R}(T_{i-1})$. 

   \item $\textit{juxtapose}(\varphi^\mathrm{tgt}, \varphi^\mathrm{prt}, \varphi^\mathrm{new})$:    Replaces the non-terminal node $\varphi^\mathrm{tgt}\in\mathcal{R}(T_{i-1})$ with the node $\varphi^\mathrm{new}$. $\varphi^\mathrm{tgt}$ takes the role of  left child of $\varphi^\mathrm{new}$ (keeping
   its descendants). The right child of  $\varphi^\mathrm{new}$ is a fresh subtree rooted at $\varphi^\mathrm{prt}$ with the new read word $w_i$ as only child.
   
\end{itemize}

Given a partial tree $T_{i-1}$, each span extended from fencepost $l-1$ to $r$ is represented according to Equation \ref{eq:aj-input} as a concatenation of (i) an embedding of the non-terminal symbol of the span ($\mathbf{c}_{l,r}$), and (ii) an embedding corresponding to the difference of the positions $\mathbf{p}_{l}$ and $\mathbf{p}_r$. All the spans of the partial tree $T_{i-1}$ are stacked together in a matrix $\mathbf{X}_i=[\mathbf{C}_i, \mathbf{P}_i]$ and then passed through a graph convolutional network (GCN) to obtain a new contextualized matrix $\tilde{\mathbf{X}}_i = [\tilde{\mathbf{C}}_i, \tilde{\mathbf{P}}_i]$, where each row  vector $\tilde{\mathbf{x}}_{l,r}$ is split as $ \tilde{\mathbf{x}}_{l,r}= [\tilde{\mathbf{c}}_{l,r}, \tilde{\mathbf{p}}_{l,r}]$ using the same input dimensions (see Equation \ref{eq:aj-input}) to separate positional from constituent information. Given the contextualization of a new input word $\overline{\mathbf{h}}_i$ with its positional embedding $\mathbf{p}_i$, the scores to select the target node $\mathbf{s}_i^\mathrm{tgt}$ are computed by two FFNs which operate with those word and span representations in the rightmost chain, denoted as $\tilde{\mathbf{X}}^\mathcal{R}_i = [\tilde{\mathbf{C}}^\mathcal{R}_i, \tilde{\mathbf{P}}^\mathcal{R}_i]$ (Equation \ref{eq:aj-target}). Finally, the scores for the parent and new nodes ($\mathbf{s}_i^\text{prt}$ and $\mathbf{s}_i^\text{new})$ are generated from $\overline{\mathbf{h}}_i$ and $\mathbf{p}_i$ vectors with the weighted representation of the rightmost chain (Equation \ref{eq:aj-prt-new}).

\begin{equation}\label{eq:aj-input}\small
    \mathbf{x}_{l,r} = [\mathbf{c}_{l,r}, (\mathbf{p}_r-\mathbf{p}_l)/2]
\end{equation}
\begin{equation}\label{eq:aj-target}\small
    \mathbf{s}^\mathrm{tgt}_i = \mathrm{FFN}_c([\tilde{\mathbf{C}}^\mathcal{R}_i, \overline{\mathbf{h}}_i]) + \mathrm{FFN}_p ([\tilde{\mathbf{P}}^\mathcal{R}_i, \mathbf{p}_i] )
\end{equation}
\begin{equation}\label{eq:aj-prt-new}\small
    \mathbf{s}_i^\mathrm{prt}, \mathbf{s}_i^\mathrm{new} = \mathrm{FFN}\Big(\big[\overline{\mathbf{h}}_i, \mathbf{p}_i,( \mathbf{s}^\mathrm{tgt}_i  \tilde{\mathbf{X}}_i^\mathcal{R})\big]\Big)
\end{equation}

Figure \ref{fig:aj-example} shows the update of a partial tree $T_3$ (Figure \ref{fig:aj-state0}) when applying the \textit{attach} (Figure \ref{fig:aj-attach}) or \textit{juxtapose} (Figure \ref{fig:aj-juxtapose}) actions. Note that the target node always belongs to the rightmost chain and at least one non-terminal node is  added at each time step, producing always a valid partial tree $T_i$.\footnote{\citet{yang2020strongly} proved that the attach-juxtapose is injective for constituent trees without unary chains.}

\section{Experiments}\label{section-experiments}

\paragraph{Setup} To create our models, we used the \texttt{supar}\footnote{\url{https://parser.yzhang.site/}} library as our starting point. It implements non-incremental parsers for the main parsing formalisms, including constituent parsing, and allows for plug-and-play integration of most large language models, including generative ones. For additional information, see Appendix \ref{ap:hyperparameters}.

\paragraph{Data} 
We use both the English Penn Treebank \cite{marcus1993building} and the set of multilingual treebanks released as a part of the SPMRL shared task \cite{seddah-etal-2013-overview}.\footnote{We do not report results for the Arabic treebank since it requires a paid license to be used.}

\paragraph{Metrics} 
We use labeled bracketing F1-score, with the \texttt{COLLINS.prm} (for PTB) and \texttt{evalb\_spmrl.prm} (for SPMRL) files.\footnote{BLOOM lacks pre-training data for German, Hungarian, Hebrew, Swedish, Polish, and Korean. As mentioned earlier, this is still useful to gather additional comprehension on how an incremental parser with a generative LLM performs on languages it was not specifically pre-trained for.}

\paragraph{Upper-bound baselines} We compare our models against counterparts that are not fully incremental. On the one hand, we consider \citet{kitaev2018constituency}'s approach as an upper-bound baseline, as it uses Transformers and a powerful CYK neural decoding method. On the other hand, we explore partially incremental versions of our strong incremental models as control baselines, where the encoder is replaced with a bidirectional encoder, specifically XLM-RoBERTa \cite{conneau2020unsupervised}.

\subsection{Results}

Table \ref{tab:zero-delay} presents the outcomes for the strict incremental models with $k = 0$, compared to the upper-bound and control parsers. The results suggest that the main challenges in competing with bidirectional systems are primarily associated with the encoder side. This finding is similar to observations made by other authors for different paradigms, such as dependency parsing, as noted by \citet{ezquerro2023challenges}. Particularly, we observe in Table \ref{tab:zero-delay} that models equipped with an incremental decoder and a non-incremental encoder (the control columns) achieve near state-of-the-art results. However, the F1-score substantially diminishes when switching to an incremental encoder. Across encoders, mGPT performs best overall. For languages not included in its pre-training data, BLOOM's performance is closer (yet usually higher) to that of the LSTM encoders, but it always performs worse than mGPT. We also observe clear differences across decoders. The transition-based decoders, while performing on average 10 points below the upper bound model, yield reasonable  representations. On the other hand, the incremental sequence-labeling decoders achieve a subpar F1 score, on average 27 points below state-of-the-art parsers and 17 points below the transition-based decoder. 

\begin{table}[tbp]
    \centering\scriptsize
    \setlength{\tabcolsep}{2.4pt}
    \renewcommand{\arraystretch}{1.7}
    \begin{tabular}{p{0.01cm}c|p{0.65cm}p{0.65cm}p{0.65cm}|p{0.65cm}p{0.65cm}p{0.65cm}|cc|c}
    \hline
    {} & {} &                                                                                                                              \multicolumn{6}{c|}{\textbf{Incremental}} &            \multicolumn{2}{c|}{\textbf{Control}} & \multirow{2}{*}{\textbf{KK}}\\[-0.5em]
    {} &                           {} &                                          \multicolumn{3}{c}{\bf SL} &                                              \multicolumn{3}{c|}{\bf TB}       &                          \bf SL &                     \bf TB & \\
    \hline
    \cvdots{pink, gptcolor,xlmcolor} & en &  40.4\SPSB{\lstm}{\mlp\abs} &  54.4\SPSB{\bloom}{\mlp\rel} & 57.4\SPSB{\mgpt}{\mlp\rel} &  77.2\SPSB{\lstm}{\gcn} & 83.5\SPSB{\bloom}{\gcn} & 85.7\SPSB{\mgpt}{\gcn} &   93.1\SPSB{\xlm}{\mlp\rel} &  94.5\SPSB{\xlm}{\gcn} & 95.5\\
    \cvdots{pink, gptcolor, xlmcolor}&  eu    &  59.0\SPSB{\lstm}{\mlp\rel} &  60.1\SPSB{\bloom}{\mlp\rel} & 64.1\SPSB{\mgpt}{\mlp\rel} &  71.4\SPSB{\lstm}{\gcn} & 76.5\SPSB{\bloom}{\gcn} & 81.8\SPSB{\mgpt}{\gcn} &   91.1\SPSB{\xlm}{\mlp\abs} &  92.8\SPSB{\xlm}{\gcn} & 93.6\\
    \cvdots{gptcolor,xlmcolor} & de    &  34.6\SPSB{\lstm}{\mlp\rel} &  46.3\SPSB{\bloom}{\mlp\abs} & 52.5\SPSB{\mgpt}{\mlp\abs} &  51.9\SPSB{\lstm}{\gcn} & 67.4\SPSB{\bloom}{\gcn} & 72.9\SPSB{\mgpt}{\gcn} &   90.7\SPSB{\xlm}{\mlp\abs} &  91.7\SPSB{\xlm}{\gcn} & 88.9\\
    \cvdots{pink, gptcolor,xlmcolor} & fr    &  39.7\SPSB{\lstm}{\mlp\abs} &  50.2\SPSB{\bloom}{\mlp\rel} & 53.8\SPSB{\mgpt}{\mlp\rel} &  64.9\SPSB{\lstm}{\gcn} & 71.7\SPSB{\bloom}{\gcn} & 74.5\SPSB{\mgpt}{\gcn} &   86.0\SPSB{\xlm}{\mlp\abs} &  86.6\SPSB{\xlm}{\gcn} & 91.5\\
    \cvdots{gptcolor,xlmcolor} & he    &  66.2\SPSB{\lstm}{\mlp\abs} &  66.4\SPSB{\bloom}{\mlp\rel} & 76.1\SPSB{\mgpt}{\mlp\rel} &  65.4\SPSB{\lstm}{\gcn} & 74.3\SPSB{\bloom}{\gcn} & 84.4\SPSB{\mgpt}{\gcn} &   91.8\SPSB{\xlm}{\mlp\abs} &  93.8\SPSB{\xlm}{\gcn} & 92.8\\
    \cvdots{gptcolor,xlmcolor} & hu    &  72.0\SPSB{\lstm}{\mlp\rel} &  69.3\SPSB{\bloom}{\mlp\rel} & 76.6\SPSB{\mgpt}{\mlp\rel} &  69.8\SPSB{\lstm}{\gcn} & 82.2\SPSB{\bloom}{\gcn} & 89.1\SPSB{\mgpt}{\gcn} &   94.5\SPSB{\xlm}{\mlp\abs} &  95.3\SPSB{\xlm}{\gcn} & 96.3\\
    \cvdots{gptcolor,xlmcolor} & ko    &  63.8\SPSB{\lstm}{\mlp\rel} &  63.7\SPSB{\bloom}{\mlp\rel} & 70.4\SPSB{\mgpt}{\mlp\rel} &  75.7\SPSB{\lstm}{\gcn} & 77.7\SPSB{\bloom}{\gcn} & 81.9\SPSB{\mgpt}{\gcn} &   89.0\SPSB{\xlm}{\mlp\rel} &  89.8\SPSB{\xlm}{\gcn} & 91.9\\
    \cvdots{gptcolor,xlmcolor} & pl    &  71.6\SPSB{\lstm}{\mlp\abs} &  71.8\SPSB{\bloom}{\mlp\rel} & 79.7\SPSB{\mgpt}{\mlp\abs} &  77.6\SPSB{\lstm}{\gcn} & 84.7\SPSB{\bloom}{\gcn} & 91.4\SPSB{\mgpt}{\gcn} &   96.2\SPSB{\xlm}{\mlp\abs} &  96.8\SPSB{\xlm}{\gcn} & 97.1\\
    \cvdots{gptcolor} & sv    &  47.6\SPSB{\lstm}{\mlp\rel} &  47.3\SPSB{\bloom}{\mlp\rel} & 60.3\SPSB{\mgpt}{\mlp\rel} &  60.4\SPSB{\lstm}{\gcn} & 64.1\SPSB{\bloom}{\gcn} & 78.2\SPSB{\mgpt}{\gcn} &   87.6\SPSB{\xlm}{\mlp\abs} &  90.2\SPSB{\xlm}{\gcn} & 92.0\\
    \hline
    \multicolumn{2}{c|}{$\mu$} &                        55.0 &                         58.8 &                       65.7 &                    68.3 &                    75.8 &                   82.2 &                       91.1 &                   92.4 & 93.3\\
    \hline
\end{tabular}

    \caption{\small Labeled F-score paired with best sequence labeling (SL) and transition-based (TB) decoder. $\mu$ represents macro average results. Superscripts denote the encoder choice:
    LSTM (\lstm), BLOOM-560M (\bloom), mGPT (\mgpt), XLM-RoBERTa (\xlm). Subscripts denote the decoder configuration: absolute (\abs), relative (\rel), GCN (\gcn) and FFN (\mlp). 
    The upper bound baseline performance (\textbf{KK}, \cite{kitaev2018constituency}) is also included. Language codes come from ISO 639-1 and left colored dots indicate the pretraining availability in LMs.}
    \label{tab:zero-delay}
\end{table}

Table \ref{tab:delay-comparison} compares our incremental models with zero delay to counterpart versions with delays of 1 and 2. The improvements are noticeable in both decoders, especially from delay zero to one. On average, for the sequence labeling decoder, moving from delay zero to one improves performance by 13.7 and 15.6 percentage points for the LSTM and mGPT encoders, respectively. Meanwhile, the improvements from delay 1 to delay 2 show clear diminishing returns, with only a 0.8 and 2.3 point improvement. The trend is similar for the transition-based decoder.  When setting $k$=$1$, it shows average improvements of 8.2 points (using the vanilla LSTM encoder) and 4.5 points (mGPT) compared to the strict incremental version. However, there is only a 0.8 and a 0.9 point improvement compared to the models with delay one. 
These diminishing returns indicate that small delays are not enough to close the gap, and strategies to improve incremental encoders such as prophecy tokens that can simulate larger delays~\cite{madureira-schlangen-2020-incremental} may be needed - although tailored for parsing and contemporary language models.

\begin{table}[tbp]
    \centering\scriptsize
    \setlength{\tabcolsep}{2.2pt}
    \renewcommand{\arraystretch}{1.7}
    \begin{tabular}{p{0.01cm}c|p{0.7cm}p{0.7cm}|p{0.7cm}p{0.7cm}|p{0.7cm}p{0.7cm}|p{0.7cm}p{0.7cm}}
\hline
    {} & {} & \multicolumn{4}{c|}{\bf SL} & \multicolumn{4}{c}{\bf TB} \\[-0.5em] 
    {} & &  \multicolumn{2}{c}{\bf LSTM (\lstm)} &  \multicolumn{2}{c|}{\bf MGPT (\mgpt)} & \multicolumn{2}{c}{\bf LSTM (\lstm)} & \multicolumn{2}{c}{\bf MGPT (\mgpt)} \\
    \hline
    \cvdots{pink,gptcolor,xlmcolor} & en   &  68.3\SB{\textcolor{boostcolor}{27.9}} &  72.4\SB{\textcolor{boostcolor}{32.0}} & 82.3\SB{\textcolor{boostcolor}{24.9}}  & 86.1\SB{\textcolor{boostcolor}{28.7}} & 83.4\SB{\textcolor{boostcolor}{6.2}}  &     84.2\SB{\textcolor{boostcolor}{7.0}} &   90.9\SB{\textcolor{boostcolor}{5.2}} &  91.6\SB{\textcolor{boostcolor}{5.9}}  \\
    \cvdots{pink,gptcolor,xlmcolor}& eu    &  78.0\SB{\textcolor{boostcolor}{19.0}} &  77.2\SB{\textcolor{boostcolor}{18.2}} & 84.8\SB{\textcolor{boostcolor}{20.7}}  & 86.8\SB{\textcolor{boostcolor}{22.7}} & 81.7\SB{\textcolor{boostcolor}{10.3}} &     81.2\SB{\textcolor{boostcolor}{9.8}} &   87.1\SB{\textcolor{boostcolor}{5.3}} &  88.0\SB{\textcolor{boostcolor}{6.2}}  \\
    \cvdots{gptcolor,xlmcolor}& de    &  57.5\SB{\textcolor{boostcolor}{22.9}} &  59.0\SB{\textcolor{boostcolor}{24.4}} & 72.8\SB{\textcolor{boostcolor}{20.3}}  & 76.4\SB{\textcolor{boostcolor}{23.9}} & 64.6\SB{\textcolor{boostcolor}{12.7}} &    64.5\SB{\textcolor{boostcolor}{12.6}} &   81.3\SB{\textcolor{boostcolor}{8.4}} & 83.3\SB{\textcolor{boostcolor}{10.4}}  \\
    \cvdots{pink,gptcolor,xlmcolor}& fr    &  59.0\SB{\textcolor{boostcolor}{19.3}} &  60.4\SB{\textcolor{boostcolor}{20.7}} & 75.2\SB{\textcolor{boostcolor}{21.4}}  & 78.5\SB{\textcolor{boostcolor}{24.7}} & 73.5\SB{\textcolor{boostcolor}{8.6}}  &    76.4\SB{\textcolor{boostcolor}{11.5}} &  81.0\SB{\textcolor{boostcolor}{6.5}} &  83.0\SB{\textcolor{boostcolor}{8.5}}  \\
    \cvdots{gptcolor,xlmcolor}& he    &   75.7\SB{\textcolor{boostcolor}{9.5}} &  76.3\SB{\textcolor{boostcolor}{10.1}} &  84.7\SB{\textcolor{boostcolor}{8.6}}  &  85.5\SB{\textcolor{boostcolor}{9.4}} & 77.7\SB{\textcolor{boostcolor}{12.3}} &    79.7\SB{\textcolor{boostcolor}{14.3}} &   87.5\SB{\textcolor{boostcolor}{3.1}} &  88.0\SB{\textcolor{boostcolor}{3.6}}  \\
    \cvdots{gptcolor,xlmcolor}& hu    &   76.4\SB{\textcolor{boostcolor}{4.4}} &   79.6\SB{\textcolor{boostcolor}{7.6}} &  84.8\SB{\textcolor{boostcolor}{8.2}}  & 87.5\SB{\textcolor{boostcolor}{10.9}} & 82.1\SB{\textcolor{boostcolor}{12.3}} &    85.2\SB{\textcolor{boostcolor}{15.4}} &   92.0\SB{\textcolor{boostcolor}{2.9}} &  92.1\SB{\textcolor{boostcolor}{3.0}}  \\
    \cvdots{gptcolor,xlmcolor}& ko    &   70.0\SB{\textcolor{boostcolor}{6.2}} &   70.0\SB{\textcolor{boostcolor}{6.2}} &  78.0\SB{\textcolor{boostcolor}{7.6}}  &  80.0\SB{\textcolor{boostcolor}{9.6}} & 77.1\SB{\textcolor{boostcolor}{1.4}}  &     77.8\SB{\textcolor{boostcolor}{2.1}} &  83.9\SB{\textcolor{boostcolor}{2.0}} &  84.6\SB{\textcolor{boostcolor}{2.7}}  \\
    \cvdots{gptcolor,xlmcolor}& pl    &  83.1\SB{\textcolor{boostcolor}{11.5}} &  82.0\SB{\textcolor{boostcolor}{10.4}} & 91.4\SB{\textcolor{boostcolor}{11.7}}  & 92.4\SB{\textcolor{boostcolor}{12.7}} & 86.2\SB{\textcolor{boostcolor}{8.6}}  &    87.8\SB{\textcolor{boostcolor}{10.2}} &  93.6\SB{\textcolor{boostcolor}{2.2}} &  94.2\SB{\textcolor{boostcolor}{2.8}}  \\
    \cvdots{gptcolor}& sv    &  64.3\SB{\textcolor{boostcolor}{16.7}} &  62.9\SB{\textcolor{boostcolor}{15.3}} & 77.3\SB{\textcolor{boostcolor}{17.0}}  & 79.1\SB{\textcolor{boostcolor}{18.8}} & 70.1\SB{\textcolor{boostcolor}{9.7}}  &     66.9\SB{\textcolor{boostcolor}{6.5}} &  82.8\SB{\textcolor{boostcolor}{4.6}} &  83.8\SB{\textcolor{boostcolor}{5.6}}  \\
    \hline
    \multicolumn{2}{c|}{$\mu$}&   63.2\SB{\textcolor{boostcolor}{13.7}} &  64.0\SB{\textcolor{boostcolor}{14.5}} & 81.3\SB{\textcolor{boostcolor}{15.6}}  &  83.6\SB{\textcolor{boostcolor}{17.9}} & 69.6\SB{\textcolor{boostcolor}{8.2}}  &     70.4\SB{\textcolor{boostcolor}{9.0}} &  86.7\SB{\textcolor{boostcolor}{4.5}} &  87.6\SB{\textcolor{boostcolor}{5.4}}  \\
    \hline
\end{tabular}

    \caption{\small LF scores with delay 1 and 2 (first and second subcolumn) Notation as in Table \ref{tab:zero-delay}. Subscripts denote performance boost over zero-delay fully incremental results from Table \ref{tab:zero-delay}.}
    \label{tab:delay-comparison}
\end{table}

\begin{figure}[!]
    \centering\small
    \begin{subfigure}[b]{0.23\textwidth}
        \label{fig:english-fscore}
        \caption{PTB}
        \includegraphics[width=\textwidth]{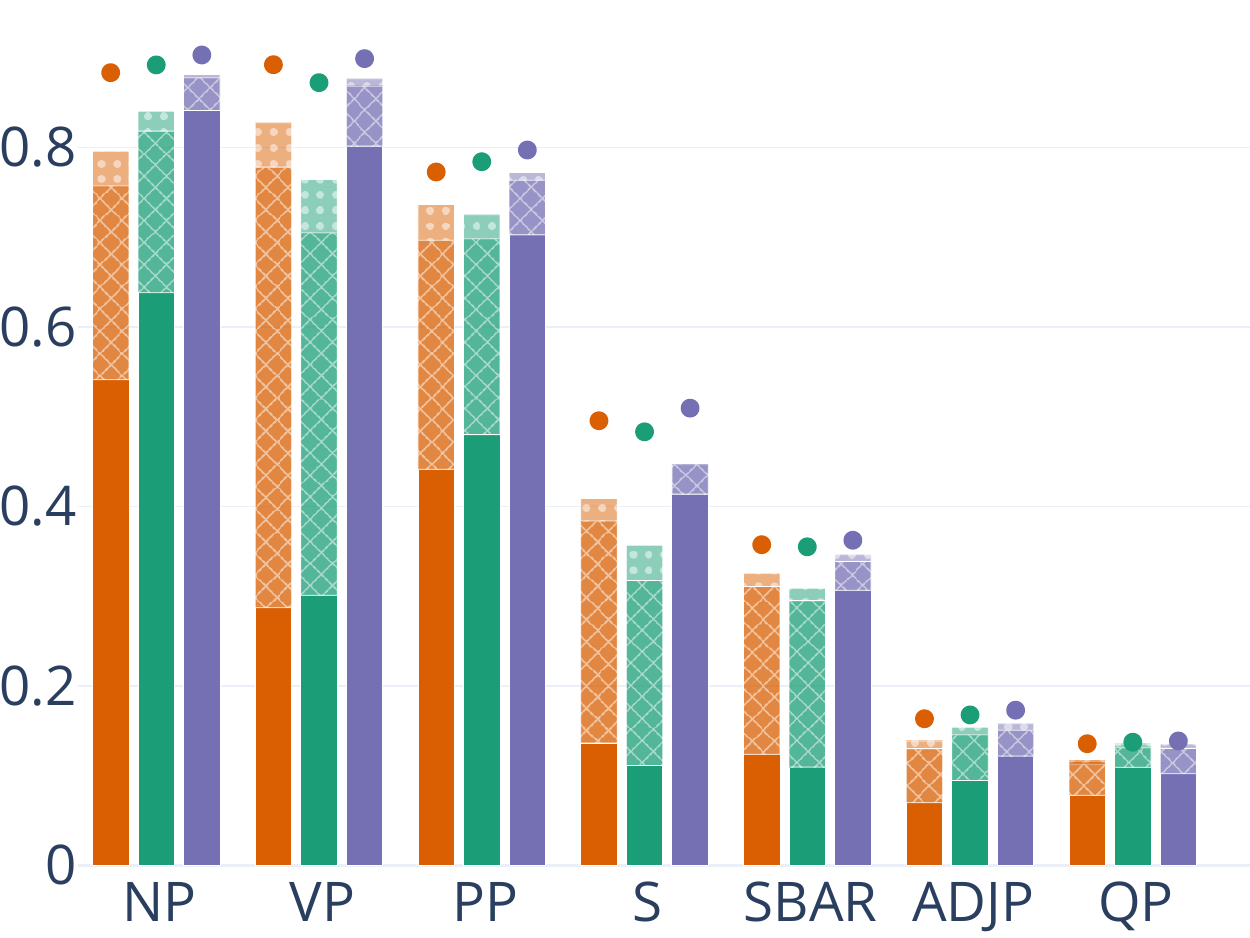}
    \end{subfigure}
    \begin{subfigure}[b]{0.23\textwidth}
        \label{fig:basque-fscore}
        \caption{Basque}
        \includegraphics[width=\textwidth]{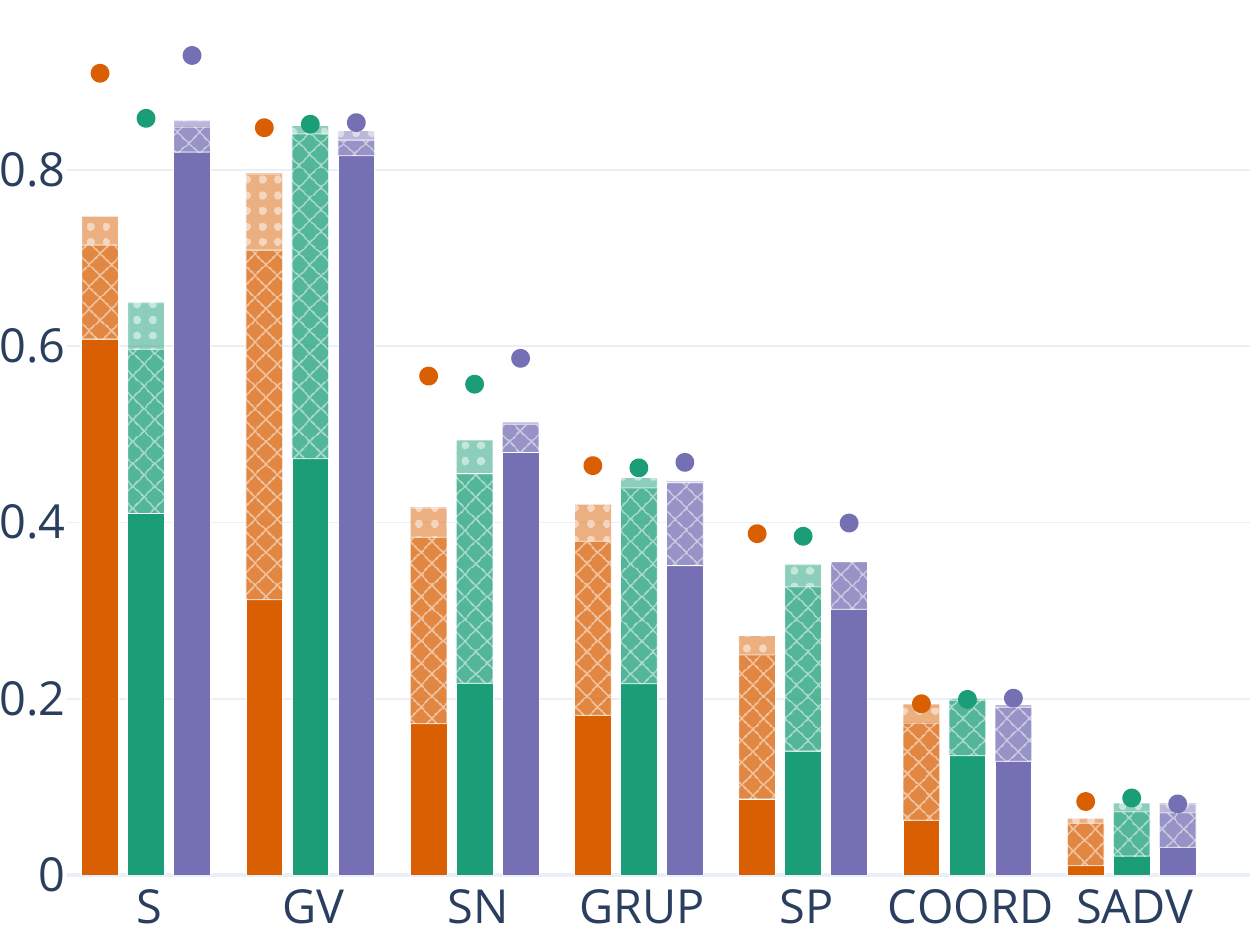}
    \end{subfigure}
    \begin{subfigure}[c]{0.23\textwidth}
        \label{fig:hebrew-fscore}
        \caption{Hebrew}
        \includegraphics[width=\textwidth]{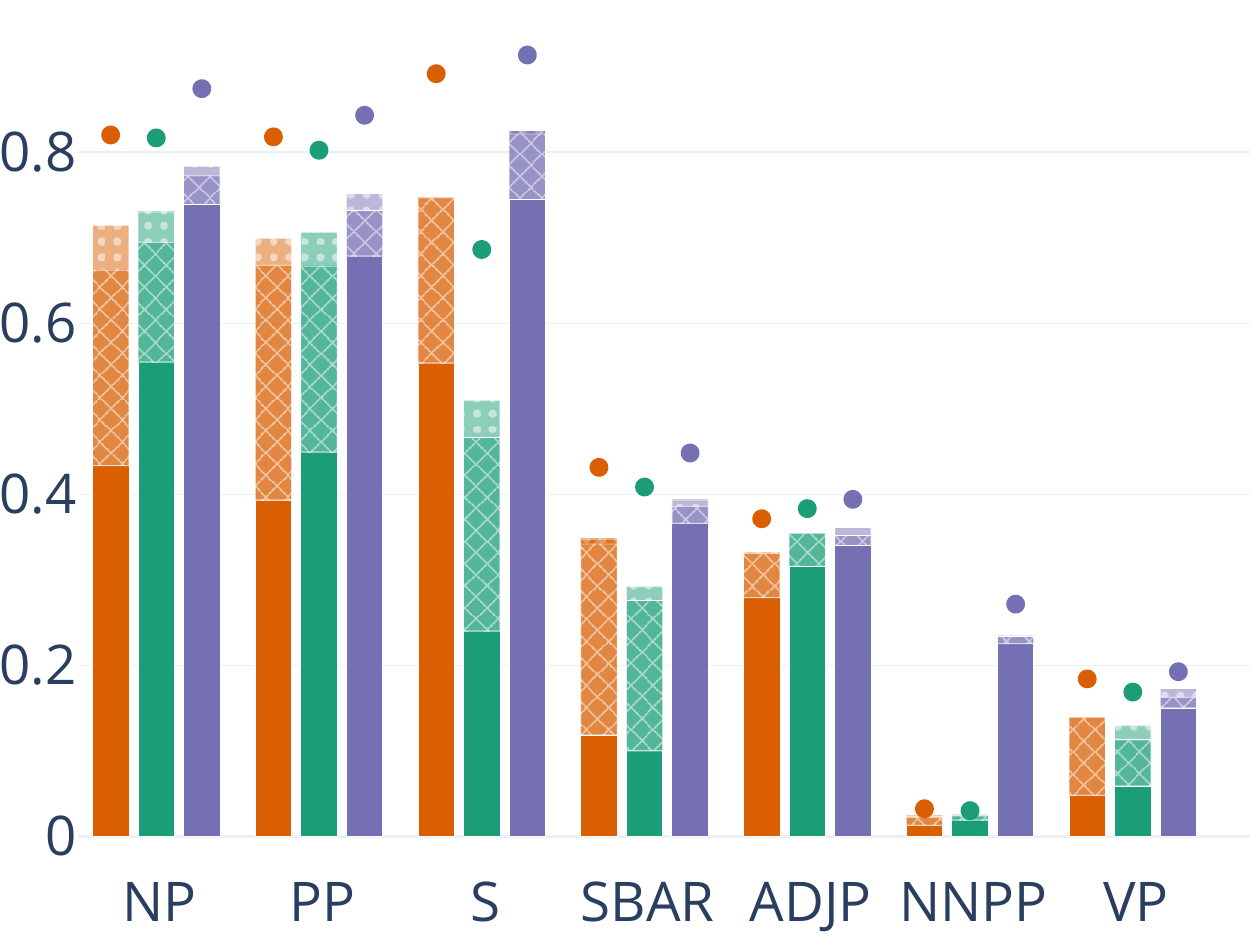}
    \end{subfigure}
    \begin{subfigure}[c]{0.23\textwidth}
        \label{fig:korean-fscore}
        \caption{Korean}
        \includegraphics[width=\textwidth]{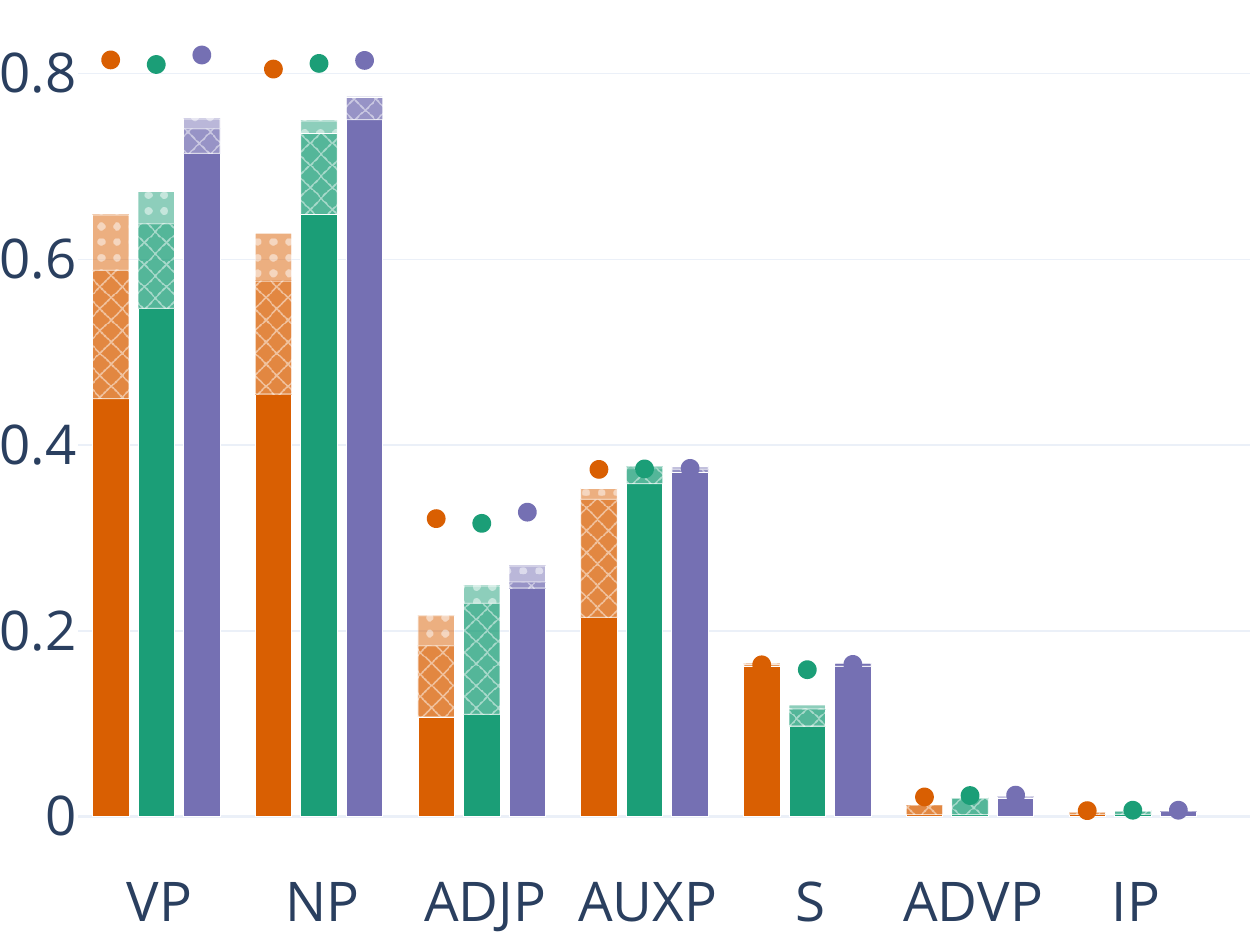}
    \end{subfigure}
    \caption{\label{fig:constituents-fscore} F-Score of absolute (orange), relative (green) and transition-based (purple) decoders with mGPT (bars) and XLM-RoBERTa (dots) encoders per constituent. Different textures are used for delay 0 (solid), 1 (dotted) and 2 (gridded).}
\end{figure}

Finally, some phrases may be more ambiguous than others in an incremental setting due to factors such as sentence structure, word order, or semantics. Figure \ref{fig:constituents-fscore} shows F1-scores for the most common non-terminals in diverse languages: English, Hebrew, Basque, and Korean, for models with $k \in [0,1,2]$. For space reasons, we include only the most coarse-grained non-terminals. Unary chains are excluded. Across the board, positive delays, especially $k=1$, have a much greater impact on sequence-labeling decoders, particularly benefiting longer span types like noun, verb, and prepositional phrases (span lengths are in Appendix Table~\ref{tab:treebank-statistics}). Also, behaviors across non-terminals and languages can vary greatly with incrementality, e.g., while delay is crucial for phrases such as VB for PTB or SBAR for Hebrew, its need is negligible for others such as Hebrew ADJP.

\section{Conclusion}

This paper introduced a set of strictly incremental encoder-decoder constituent parsers, using generative language models and two types of decoders: one based on parsing as tagging, and the other on transition-based parsing with partial graph neural network representations. 
We tested the models in a diverse multilingual setting and also simulated human reading processes with positive delays of a few upcoming words. The results suggest that a significant portion of future challenges may be centered on the encoding side, and in how different phrases might be affected by the absence of bidirectionality. In this context, exploring research lines to inform the decoder, such as speculative real-time generation of next tokens in real time, could be a valuable step to explore parsing methods closer to human reading processes.

\section*{Acknowledgments}
We thank the anonymous reviewers for their very complete and useful suggestions.
This work has received funding by the European Research Council (ERC), under the Horizon Europe research and innovation programme (SALSA, grant agreement No 101100615), ERDF/MICINN-AEI (SCANNER-UDC, PID2020-113230RB-C21), Xunta de Galicia (ED431C 2020/11), Grant GAP (PID2022-139308OA-I00) funded by MCIN/AEI/10.13039/501100011033/ and by ERDF “A way of making Europe”, Cátedra CICAS (Sngular, University of A Coruña), and Centro de Investigación de Galicia ‘‘CITIC’’, funded by the Xunta de Galicia through the collaboration agreement between the Consellería de Cultura, Educación, Formación Profesional e Universidades and the Galician universities for the reinforcement of the research centres of the Galician University System (CIGUS).

\section*{Limitations}

\paragraph{Non-monotonicity} Our definition of incrementality applies exclusively to monotonic parsers. In cases of non-monotonicity, a parser might abandon its existing partial output and revise it as new information comes in. This is carefully discussed in \cite{ezquerro2023challenges} for dependency parsing. Similarly, we chose to focus solely on monotonic constituent parsers. First, our goal is to maintain a straightforward implementation that is on par with others, avoiding the added complexity that repair strategies entail. Second, dealing with non-monotonicity requires thinking a thorough evaluation framework. In this respect, comparing against (partial) incremental parsers is challenging, as metrics must account for partial analysis, which is not accommodated by the standard bracketing F1-measure. In turn, such metrics on partial analyses are meaningless for non-incremental parsers, which often do not even produce any partial outputs, precluding direct comparison against them.

\paragraph{Discontinuous constituent parsing} We restricted our analysis to continuous constituent parsing and observed that modern incremental parsers still exhibit shortcomings in this area. Studying the impact of incrementality on discontinuities, i.e., discontinuous spans within a sentence that form specific constituents, presents a more challenging aspect of constituent parsing. This phenomenon is particularly observed in languages with free word order. In this regard, there are several avenues to explore. For example, we could draw inspiration from the incremental transition-based algorithm described by \citet{coavoux-crabbe-2017-incremental}, or the sequence labeling approach suggested by \citet{vilares-gomez-rodriguez-2020-discontinuous}, which shows potential to be adapted to an incremental setup.

\paragraph{Experiments on lower-resourced languages} Unlike in other paradigms like dependency parsing, the availability of a diverse range of treebanks spanning various typologies is more limited for constituent parsing. We used the treebanks presently at our disposal, which include the  English Penn Treebank and the SPMRL treebanks. However, it is worth noting that we were unable to access the Arabic dataset due to its paid license. Yet, our experiments consider: English (Indo-European, Germanic), Basque, German (Indo-European, Germanic), French (Indo-European, Romance), Hebrew (Afro-Asiatic, Semitic), Hungarian (Uralic, Ugric), Korean, Polish (Indo-European, Slavic) and Swedish (Indo-European, Germanic).

\paragraph{Availability  of multilingual large language models} Research on generative LLMs is extensive, and many models are being released contemporaneously with this paper. However, highly multilingual versions are more rare. Two main available resources are BLOOM and mGPT, both of which we evaluated. This double evaluation allowed us to establish differences in performance, particularly in terms of incrementality, depending on whether the models contained pre-training data for a given target language or not.

\paragraph{Computational capabilities} We lacked access to extensive computing infrastructure or a budget for cloud-based scaling that would allow us to fine-tune more powerful multilingual language models such as the LLaMa family.  We had access to NVIDIA GeForce RTX 3090 servers (each with 24GB) and one NVIDIA A100 GPU (with 80GB). We managed to fine-tune the smallest BLOOM language model (560M parameters) within reasonable time frames. Although we could technically fit the 3B version on the A100 GPU with a minimal batch size, the impractical training duration made it infeasible for a comprehensive multilingual study like the one proposed in this work.

\bibliography{custom}
\bibliographystyle{acl_natbib}

\appendix
\section{Appendix}
\label{sec:appendix}
\subsection{Treebank statistics}
Table \ref{tab:treebank-statistics} shows the frequency and average length (defined as the difference between initial and final fencepost) of the constituents displayed in Figure \ref{fig:constituents-fscore}. 

\begin{table}[h]
    \centering\scriptsize
    \setlength{\tabcolsep}{2pt}
    \renewcommand{\arraystretch}{1.3}
    \begin{tabular}{c|ccccccc}
    \hline
    \textbf{en} &     NP &      VP &     PP &      S &   SBAR &   ADJP &         QP \\
    \cline{2-8}
    $\lambda$   &   4.96 &    11.2 &   5.96 &  12.71 &   12.2 &   5.02 &       2.97 \\
    \%          &  43.13 &   24.56 &  16.95 &   6.29 &    3.6 &   1.79 &        1.7 \\
    \hline 
    \textbf{eu} &      S &      GV &     SN &   GRUP &     SP &  COORD &       SADV \\
    \cline{2-8}
    $\lambda$   &   8.86 &    2.31 &   3.66 &   2.26 &   3.38 &   2.04 &       2.61 \\
    \%          &  34.07 &   18.47 &  15.46 &  12.38 &   11.2 &    2.8 &       2.08 \\
    \hline
    \textbf{de} &     NP &      PP &     VN &   SENT &  COORD &  VPINF &         NC \\
    \cline{2-8}
    $\lambda$   &   6.37 &     6.5 &   2.52 &  30.44 &    9.4 &  12.35 &       2.59 \\
    \%          &  32.79 &   26.25 &   7.04 &   6.04 &   4.53 &    3.2 &       2.92 \\
    \hline
    \textbf{fr} &     NP &      PP &      S &     VP &     AP &     PN &        CNP \\
    \cline{2-8}
    $\lambda$   &   3.42 &    3.69 &    8.5 &   5.81 &   3.09 &   2.16 &       4.72 \\
    \%          &  28.66 &   25.55 &  21.22 &   9.16 &   4.37 &   2.98 &       2.75 \\
    \hline
    \textbf{he} &     NP &      PP &      S &   SBAR &   ADJP &   NNPP &         VP \\
    \cline{2-8}
    $\lambda$   &   5.14 &    5.93 &  17.21 &  13.15 &   2.69 &   2.65 &      10.67 \\
    \%          &  44.19 &   21.68 &  15.21 &    5.4 &   4.44 &    3.2 &       2.56 \\
    \hline
    \textbf{hu} &     NP &      CP &   ADJP &     PP &     XP &   ADVP &          V \\
    \cline{2-8}
    $\lambda$   &   3.58 &   14.97 &   3.79 &   4.09 &   7.36 &   2.88 &        2.0 \\
    \%          &  57.11 &    29.3 &   7.54 &   3.97 &   1.26 &   0.66 &       0.16 \\
    \hline
    \textbf{ko} &     VP &      NP &   ADJP &   AUXP &      S &   ADVP &         IP \\
    \cline{2-8}
    $\lambda$   &   6.76 &    3.79 &   6.29 &   2.14 &  13.45 &   3.03 &       2.06 \\
    \%          &  49.75 &    36.0 &   8.04 &    4.1 &   1.53 &   0.39 &       0.16 \\
    \hline
    \textbf{pl} &    FNO &  ZDANIE &    FPM &    FWE &    FZD &    FPT &  FORMACZAS \\
    \cline{2-8}
    $\lambda$   &   3.61 &    8.09 &    3.3 &   3.57 &    8.4 &    3.9 &       2.04 \\
    \%          &  34.18 &   28.15 &  19.79 &   5.39 &   3.56 &   3.34 &       2.21 \\
    \hline
    \textbf{sv} &     NP &       S &     PP &     VP &     XP &     AP &        AVP \\
    \cline{2-8}
    $\lambda$   &   4.61 &   12.14 &   4.61 &   7.64 &   4.24 &   2.65 &       3.51 \\
    \%          &   31.1 &   27.35 &  20.69 &  10.35 &   6.46 &   2.48 &       0.88 \\
    \hline
    \end{tabular}
    \caption{\label{tab:treebank-statistics} Frequency (\%) and average length ($\lambda$) of most frequent constituents of each treebank. Root and unary spans were removed.}
\end{table}

\subsection{Hyperparameters configuration}\label{ap:hyperparameters}
    Tables \ref{tab:model-hyp} and \ref{tab:train-hyp} show the configuration of the models and the training hyperparameters for each encoder type. In pretrained models, each sentence $(w_1,...,w_n)$ was passed through all encoder layers to compute the last hidden state $(\mathbf{e}_1, ..., \mathbf{e}_n)$ and then projected to a new reduced space of dimension $h$ with a feed-forward network. 
    The final reduced sequence $\mathbf{H}=(\mathbf{h}_1,...,\mathbf{h}_n)$ is the one passed to the delay module. In the case of non-pretrained encoders, each word $w_i$ was represented as a concatenation of (i) a word embedding of dimension $h_w$, (ii) the PoS tag embedding of dimension $h_p$ and (iii) the last hidden state of a Character-LSTM \cite{dozat2017deep} of dimension $h_c$, resulting into a final input embedding $\mathbf{w}_i\in\mathbb{R}^{h_w+h_p+h_c}$. The input matrix $\mathbf{W}=(\mathbf{w}_1,..,\mathbf{w}_n)$  is introduced to the LSTM encoder (with randomly initialized weights) and its last hidden states $(\mathbf{h}_1,...,\mathbf{h}_n)$ are passed through the delay module and the decoder. The decoder is a 3-layered Graph Convolutional Layer for the Attach-Juxtapose parser or a feed-forward network for the case of the sequence labeling decoder. The complete network was trained with the CrossEntropy loss function and AdamW as optimizer, adapting the batches to the model size. Dropout was set in both encoder and decoder and the best validation performance was finally retrieved.

\begin{table}[h]
    \centering\scriptsize
    \setlength{\tabcolsep}{3pt}
    \begin{tabular}{c|ccc|cc}
         \multirow{2}{*}{\textbf{Hyp.}} & \multicolumn{3}{c|}{\textbf{LLM}} & \multicolumn{2}{c}{\textbf{Non-pretrained}} \\
         & XLM\SP{\xlm} & BLOOM\SP{\bloom} & mGPT\SP{\mgpt} & LSTM\SP{\lstm} & BiLSTM\SP{\bilstm}  \\
         \hline 
         word emb. ($h_w$) & \multicolumn{3}{c|}{-} & \multicolumn{2}{c}{300} \\ 
         PoS emb. ($h_p$) & \multicolumn{3}{c|}{-} & \multicolumn{2}{c}{100} \\ 
         char. emb. & \multicolumn{3}{c|}{-} & \multicolumn{2}{c}{50} \\ 
         char. LSTM ($h_c$) &  \multicolumn{3}{c|}{-} & \multicolumn{2}{c}{100} \\ 
         \# enc. layers & \multicolumn{3}{c|}{1} & \multicolumn{2}{c}{4} \\ 
         enc. emb. ($h$) & \multicolumn{3}{c|}{100} & \multicolumn{2}{c}{400} \\ 
         \% enc. dropout &  \multicolumn{3}{c|}{0.33} & \multicolumn{2}{c}{0.33\SB{sh.}} \\ 
         \# GCN layers & \multicolumn{3}{c|}{3} & \multicolumn{2}{c}{3} \\ 
         \# FFN layers &  \multicolumn{3}{c|}{1} & \multicolumn{2}{c}{1} \\ 
         \% dec. dropout &  \multicolumn{3}{c|}{0.33} & \multicolumn{2}{c}{0.33\SB{sh.}} \\ 
    \end{tabular}
    \caption{\label{tab:model-hyp}Model configuration for pretrained and non-pretrained models. The number of encoder layers for LLMs refers to the number of last hidden states obtained for each word. LSTM-based encoders use the shared-dropout technique \citep{yarin2016dropout} as described in \citet{dozat2017deep}.}
\end{table}

\begin{table}[h]
    \centering\scriptsize
    \setlength{\tabcolsep}{4pt}
    \begin{tabular}{c|ccc|cc}
         \multirow{2}{*}{\textbf{Hyp.}} & \multicolumn{3}{c|}{\textbf{LLM}} & \multicolumn{2}{c}{\textbf{Non-pretrained}} \\
         & XLM\SP{\xlm} & BLOOM\SP{\bloom} & mGPT\SP{\mgpt} & LSTM\SP{\lstm} & BiLSTM\SP{\bilstm}  \\
         \hline 
         optimizer & \multicolumn{3}{c|}{AdamW} & \multicolumn{2}{c}{AdamW} \\ 
         lr & \multicolumn{3}{c|}{5e-5} & \multicolumn{2}{c}{1e-3} \\ 
         lr decay & \multicolumn{3}{c|}{linear (0.5)} & \multicolumn{2}{c}{exponential (0.1)} \\ 
         epochs & \multicolumn{3}{c|}{30} & \multicolumn{2}{c}{200} \\ 
         batch size & 500 & 500 & 100 & \multicolumn{2}{c}{5000}
    \end{tabular}
    \caption{\label{tab:train-hyp}Training hyperparameters for pretrained and non-pretrained models. AdamW is set as optimizer with $\beta_0=0.9$, $\beta_1=0.9$ and $\varepsilon=10^{-12}$, and batch sampling is fixed to minimize sequence padding.}
\end{table}

Finally, Table \ref{tab:exectime} displays various estimates of inference speeds for different models.

\begin{table}[hbtp!]
    \centering\scriptsize
    \renewcommand{\arraystretch}{1.1}
    \begin{tabular}{c|ccccc|}
         & \multicolumn{5}{c|}{\bf SL (\mlp)} \\
         \hline 
         & LSTM\SP{\lstm} & BiLSTM\SP{\bilstm} &  BLOOM\SP{\bloom} & mGPT\SP{\mgpt} & XLM\SP{\xlm}\\
         \hline 
            en  & 856.02 & 688.98 & 354.21 & 144.03 & 403.09 \\
            eu  & 2096.61 & 1265.82 & 424.77 & 168.95 & 523.03 \\
            fr  & 678.24 & 467.33 & 281.33 & 116.02 & 313.66 \\
            de  & 1348.32 & 969.85 & 306.75 & 155.42 & 417.9 \\
            he  & 497.8 & 673.3 & 192.07 & 122.14 & 323.24 \\
            hu  & 1258.99 & 809.41 & 233.1 & 124.26 & 390.06 \\
            ko  & 1838.16 & 1477.9 & 261.41 & 151.59 & 486.57 \\
            pl  & 2091.04 & 1725.57 & 439.35 & 242.28 & 631.53 \\
            sw  & 1465.52 & 1132.12 & 325.1 & 174.44 & 507.27 \\
            \hline 
            $\mu$ & 1347.86 & 1023.36 & 313.12 & 155.46 & 444.04 \\
         \multicolumn{6}{c}{}\\
          & \multicolumn{5}{c|}{\bf TB (\gcn)} \\
         \hline 
         & LSTM\SP{\lstm} & BiLSTM\SP{\bilstm} &  BLOOM\SP{\bloom} & mGPT\SP{\mgpt} & XLM\SP{\xlm}\\
         \hline 
        en &  191.21  &  188.75   &    146.69   &   90.83   &    152.34   \\
        eu &  569.21  &  524.83   &   295.27   &  135.45   &   341.40   \\
        fr &  109.87  &  106.04   &    90.20   &   62.09   &    94.29   \\
        de &   249.92  &  241.20   &   160.34   &   102.48   &   190.38   \\
        he &  164.55  &  155.95   &     94.11   &   72.38   &   130.94   \\
        hu &  235.16  &  224.06   &   138.12   &   90.26   &   181.44   \\
        ko &  570.62  &  517.57   &   200.15   &  128.72   &   306.19   \\
        pl &  717.30  &   586.34   &   297.53   &  196.40   &    400.62   \\
        sw &  331.67  &  286.10   &   196.45   &  116.33   &     229.96   \\
        \hline 
        $\mu$ & 348.83 & 314.54 & 179.87 & 110.55 & 225.28 \\
    \end{tabular}
    \caption{\label{tab:exectime}Inference speed (in sentences per second) of the evaluated models across different languages. Symbols come from Table \ref{tab:zero-delay}.}
\end{table}

\end{document}